\pdfoutput=1
\documentclass[11pt]{article}
\usepackage{acl}
\usepackage{times}
\usepackage{latexsym}
\usepackage[T1]{fontenc}
\usepackage[utf8]{inputenc}
\usepackage{microtype}
\usepackage{amsmath,bm}
\usepackage{diagbox}
\usepackage{multirow}
\usepackage[symbol]{footmisc}
\usepackage{hyperref}

\usepackage{bbding}
\usepackage{booktabs}
\usepackage{amssymb}

\usepackage{graphicx} 
\usepackage{float} 
\usepackage{subfigure} 

\title{Incorporating Dynamic Semantics into Pre-Trained Language Model for Aspect-based Sentiment Analysis}

\author{Kai Zhang$^{1}$, Kun Zhang$^{2}$, Mengdi Zhang$^{3}$, Hongke Zhao$^{4}$, Qi Liu$^{1,}$\thanks{\ \ \ Corresponding author.}\\ {\bf Wei Wu}$^{3}$, {\bf Enhong Chen}$^{1}$\\

$^{1}$ School of Data Science, University of Science and Technology of China \\
$^{2}$ School of Computer Science and Information Engineering, Hefei University of Technology \\
$^{3}$ Meituan; 
$^{4}$ College of Management and Economics, Tianjin University \\

\texttt{kkzhang0808@mail.ustc.edu.cn}; 
\texttt{\{qiliuql,cheneh\}@ustc.edu.cn} \\ 
\texttt{\{zhang1028kun,wuwei19850318,mdz\}@gmail.com}; 
\texttt{{hongke}@tju.edu.cn}
 }

\begin{document}
\maketitle

\begin{abstract}
Aspect-based sentiment analysis (ABSA) predicts sentiment polarity towards a specific aspect in the given sentence. 
While pre-trained language models such as BERT have achieved great success, incorporating dynamic semantic changes into ABSA remains challenging.
To this end, in this paper, we propose to address this problem by  \underline{D}ynamic \underline{R}e-weighting \underline{BERT} (\textbf{DR-BERT}), a novel method designed to learn dynamic aspect-oriented semantics for ABSA. Specifically, we first take the {Stack-BERT layers} as a primary encoder to grasp the overall semantic of the sentence and then fine-tune it by incorporating a lightweight \underline{D}ynamic \underline{R}e-weighting \underline{A}dapter (DRA). Note that the DRA can pay close attention to a small region of the sentences at each step and re-weigh the vitally important words for better aspect-aware sentiment understanding.
Finally, experimental results on three benchmark datasets demonstrate the effectiveness and the rationality of our proposed model and provide good interpretable insights for future semantic modeling.
\end{abstract}

\section{Introduction}
\label{intro}
Aspect-based sentiment analysis is a branch of sentiment analysis, which aims to identify sentiment polarity of the specific aspect in a sentence~\cite{jiang2011target}. For example, given a sentence \emph{``The restaurant has attentive service, but the food is terrible.''}, the task aims to predict the sentiment polarities towards \emph{``service''} and  \emph{``food''}, which should be positive and negative respectively.

As a fundamental technology, the ABSA task has broad applications, such as recommender system~\cite{chin2018anr,zhang2021sifn} and question answering~\cite{wang2019aspect}. 
Therefore, a great amount of research has been attracted from both academia and industry. Among them, deep neural networks (DNN)~\cite{nguyen2015phrasernn,tang2015effective,tang2016aspect,zheng2020replicate}, attention mechanism~\cite{wang2016attention,ma2017interactive} and graph neural/attention networks~\cite{huang2019syntax,zhang2019aspect,wang2020relational} have significantly improved the performance through deep feature alignment between the aspect representations and context representations.

Recently, the large-scaled pre-trained language models, such as Bidirectional Encoder Representations from Transformers (BERT)~\cite{devlin2019bert}, realize a breakthrough for improving many language tasks, which further attracts considerable attention to enhance the semantic representations. In ABSA, \citet{xu2019bert} designed BERT-PT, which explores a novel post-training approach on the BERT model. \citet{song2019attentional} further proposed a text pair classification model BERT-SPC, which prepares the input sequence by appending the aspects into the contextual sentence.
Although great success has been achieved by the above studies, some critical problems remain when directly applying attention mechanisms or fine-tuning the pre-trained BERT in the task of ABSA.

Specifically, most of the existing approaches select all the important words from a contextual sentence at one time. However, according to neuroscience studies, the essential words during semantic comprehension are dynamically changing with the reading process and should be repeatedly considered~\cite{kuperberg2007neural,tononi2008consciousness,brouwer2021neurobehavioral}.  For example, when judging the sentiment polarity of the aspect \emph{``system memory''} in a review sentence \emph{``It could be a perfect laptop if it would have faster system memory and its radeon would have DDR5 instead of DDR3''}, the important words should change from general sentiment words \{\emph{``faster'', ``perfect'', ``laptop''}\} into aspect-aware words \{\emph{``would have'', ``faster'', ``could'', ``be'', ``perfect''}\}. Through these dynamic changes, the sentiment polarity will change from {positive} to the ground truth sentiment label {negative}. 

Meanwhile, simply initializing the encoder with a pre-trained BERT does not effectively boost the performance in ABSA as we expected~\cite{huang2019syntax,xu2019bert,wang2020relational}. One possible reason could be that training on two specific tasks, i.e., Next Sentence Prediction and Masked LM, with rich resources leads to better semantic of the overall sentences. However, the ABSA task is conditional, which means the model needs to understand the regional semantics of sentences by fully considering the given aspect. For instance, BERT tends to understand the global sentiment of the above sentence \emph{``It could be a perfect laptop ... of DDR3''} regardless of which aspect is given. But in ABSA, the sentence is more likely to be different sentiment meanings for different aspects (e.g., {negative} for \emph{``system memory''} while {positive} for \emph{``DDR5''}). Therefore, the vanilla BERT is hardly to pay closer attention to relevant information for the specific aspect, especially when there are multiple aspects in one sentence. 

To equip the pre-trained models with the ability to capture the aspect-aware dynamic semantics, we present a {\underline{D}ynamic \underline{R}e-weighting \underline{BERT}} (\textbf{DR-BERT}) model, which considers the aspect-aware dynamic semantics in a pre-trained learning framework.  Specifically, we first take the {Stack-BERT layers} as primary sentence encoder to learn overall semantics of the whole sentences. 
Then, we devise a \underline{D}ynamic \underline{R}e-weighting \underline{A}dapter (DRA), which aims to pay most careful attention to a small region of the contextual sentence and dynamically select and re-weight one critical word at each step for better aspect-aware sentiment understanding. 
Finally, to overcome the limitation of vanilla BERT mentioned above, we incorporate the light-weighted DRA into each BERT encoder layer and fine-tune it to adapt to the ABSA task. We conduct extensive experiments on three widely-used  datasets where the results demonstrate the effectiveness, rationality and interpretability of the proposed model.

\begin{figure*}
	\centering
	\includegraphics[width=1.55\columnwidth]{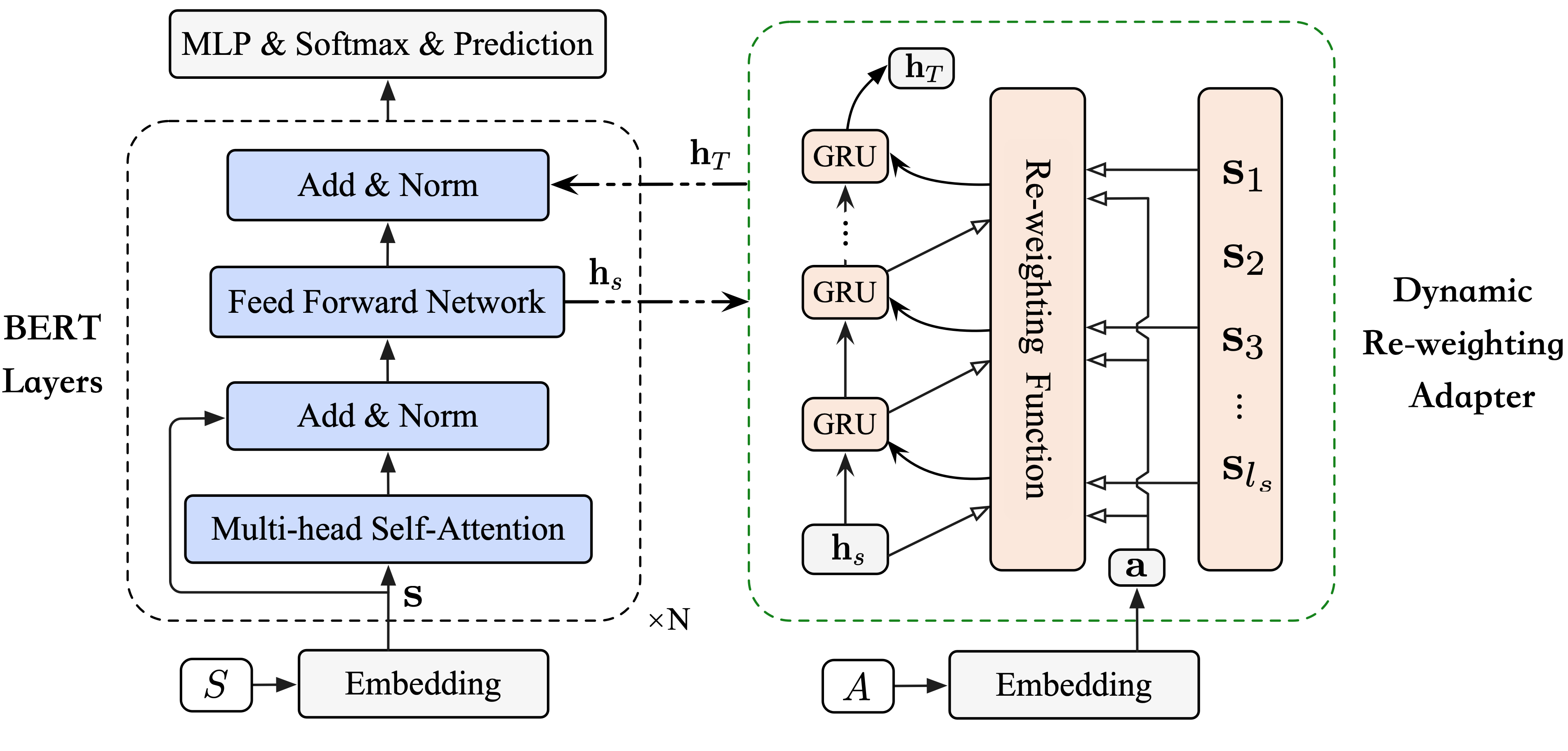}
	\vspace{-0.25cm}
	\caption{An illustration of the proposed framework. The blue blocks constitute a pre-trained BERT model which are frozen during fine-tuning, and the right block represents the dynamic re-weighting adapter that is inserted after each BERT encoder layer and trained during fine-tuning. Moreover, $S$ and $A$ represent the sentence sequence and the aspect sequence respectively. $N$ indicates the number of layers of the BERT encoder.
	}
	\label{fig:model}
	\vspace{-0.2cm}
\end{figure*}

\section{Related Work}
\vspace{-0.05cm}
\subsection{Aspect-based Sentiment Analysis} 
\vspace{-0.08cm}
\label{absa}
Aspect-based sentiment analysis identifies specific aspect's sentiment polarity in the sentence. Some approaches~\cite{ding2007utility,jiang2011target,kiritchenko2014nrc} designed numerous rules-based models for ABSA. For example, \citet{ding2007utility} first performed dependency parsing to determine sentiment polarity about the aspects. 

In recent years, most research studies make use of the attention mechanism to learn the word's semantic relation~\cite{tang2015effective,tang2016aspect,wang2016attention,ma2017interactive,xing2019earlier,liang2019novel,zhang2021eatn}. Among them, 
\citet{wang2016attention} proposed an attention-based LSTM to identify important information relating to the aspect. \citet{ma2017interactive} developed an interactive attention to model the aspect and sentence interactively. \citet{fan2018multi} deﬁned a multi-grained network to link the words from aspect and sentence. \citet{li2018transformation} designed a target-specific network to integrate aspect information into sentence. \citet{tan2019recognizing} introduced a dual attention to distinguish conflicting opinions.

In addition, another research trend is to leverage syntactic knowledge to learn syntax-aware features of the aspect~\cite{tang2019progressive,huang2019syntax,zhang2019aspect,sun2019utilizing,wang2020relational,tang2020dependency,chen2020inducing,li2021dual,tian2021aspect}. For example, \citet{tang2020dependency} developed dependency graph enhanced dual-transformer network to fuse the ﬂat representations.
More recently, pre-trained methods have been proved remarkably successful in the ABSA task. \citet{song2019attentional} devised an attentional encoder and a BERT-SPC model to learn features between aspect and context. \citet{wang2020relational} reshaped the dependency trees and proposed a relational graph attention network to encode the syntax relation feature. \citet{tian2021aspect} explicitly utilize dependency types with a type-aware graph networks to learn aspect-aware relations.

However, these methods largely ignore the procedure of dynamic semantic comprehension~\cite{kuperberg2007neural,kuperberg2016we,wang2017dynamic,zhang2019dynamic,brouwer2021neurobehavioral} and can not fully reveal dynamic semantic changes of the aspect-related words. Thus, it's hard for ABSA models to achieve the same performance as human-level sentiment understanding. 

\subsection{Human Semantic Comprehension} 
\vspace{-0.05cm}
Actually, no matter in the early days or now, imitating the procedure of human semantic comprehension has always been one of the original intention of many studies~\cite{bezdek1992relationship,wang2017dynamic,zheng2019human,li2019teach,zhang2019drr,peng2020bi,golan2020controversial}, such as machine reading comprehension~\cite{zhang2019drr,peng2020bi}, visual object detecting~\cite{spampinato2017deep} and relevance estimation~\cite{li2019teach}.
For example, attention mechanism~\cite{vaswani2017attention} has a widespread influence, which allows the model to focus on important parts of the input as human’s attention. 
\citet{spampinato2017deep} aimed to learn human–based features via brain-based visual object. \citet{wang2017dynamic} built a dynamic attention model to model human preferences for article recommendation. 

Moreover, some psychologists and psycholinguists have also done many research on the mechanisms of human semantic comprehension~\cite{kuperberg2007neural,kuperberg2016we,brouwer2021neurobehavioral}. Specifically, some scholars~\cite{yang1999reading,rayner1998eye} found that most people may focus on 1.5 words. Moreover, \citet{koch2007attention} and \citet{tononi2008consciousness} assumed that people can only remember the meaning of about 7 to 9 words at each time. The phenomenons indicate that most people only focused on a small region of the sentence at one time and need to repeatedly process important parts for better semantic understanding~\cite{sharmin2015dynamic}. 

Inspired by the above research and linguistic psychology theories, in this paper, we explore aspect-aware semantic changes of the ABSA task by incorporating the procedure of dynamic semantic comprehension into the pre-trained language model.

\section{Dynamic Re-weighting BERT}
\vspace{-0.1cm}
In this section, we introduce the technical detail of DR-BERT. Specifically, we start with the problem definition, followed by an overall architecture of DR-BERT as illustrated in Figure~\ref{fig:model}.

\vspace{0.08cm}
\noindent \textbf{Problem Definition}\ \ \ In ABSA, a sentence-aspect pair ($S, A$) is given. In this paper, the sentence is represented as ${S} = \{w_1^s, w_2^s, ..., w_{l_s}^s\}$ which consists of a series of $l_s$ words. The specific aspect is denoted as ${A} = \{w_{1}^a, w_{2}^a, ..., w_{l_a}^a\}$ which is a part of ${S}$. $l_a$ is the length of aspect words. The goal of ABSA is to learn a sentiment classifier that can precisely predict the sentiment polarity of sentence ${S}$ for specific aspect ${A}$. As the aspect-related information plays a key role in the prediction~\cite{li2018transformation,zheng2020replicate}, this paper aims to dynamically select and encode the aspect-aware semantic information through the proposed model.

\vspace{0.08cm}
\noindent \textbf{Overall Architecture}\ \ \ DR-BERT mainly contains two components (i.e., BERT encoder and Dynamic Re-weighting Adapter), together with two modules (i.e., the embedding module and sentiment prediction module). The technical details of each part will be elaborated on as follows.

\vspace{-0.08cm}
\subsection{Embedding Module} 
\vspace{-0.08cm}
To represent semantic information of the aspect words and context words better, we first map each word into a low-dimensional vector. Specifically, the inputs of DR-BERT are the sentence sequence and the corresponding aspect sequence. 
For the sentence sequence, we construct the BERT input as ``[CLS]'' + sentence +``[SEP]'' and the sentence ${S} = \{w_1^s, w_2^s, ..., w_{l_s}\}$ can be transformed into the hidden states $\mathbf{s} = \left\{\mathbf{s}_{i} \mid i=1,2, \ldots, l_{s}\right\}$ with BERT embedding. For aspect sequences, we adopt the same method to get the representation vector of each word. Thus, through the embedding module, the aspect sequence ${A} = \{w_{1}^a, w_{2}^a, ..., w_{l_a}^a\}$ is mapped to $\mathbf{a}^s = \left\{\mathbf{a}_{j} \mid j=1,2, \ldots, l_{a}\right\}$. Note that, if the aspect sequence is a single word like \emph{``food''}, the aspect representation is the embedding of the single word \emph{``food''}. While for the cases where the sequence contains multiple words such as \emph{``system memory''}, the aspect representation is the average of each word embedding~\cite{sun2015modeling}. We can denote the aspect embedding process as: 
\begin{equation}
	\mathbf{a} =  
	\left\{
             \begin{array}{lr}
             \mathbf{a}_1 ,  \ \ \mathrm{if} \ \  {l_{a}=1}\ , &\\
              \\
             ({\sum_{j=1}^{l_{a}} {\mathbf{a}_j}}) / \ { l_{a}}\ ,  \ \  \mathrm{if} \ \  {l_{a}>1}\ , 
             \end{array}
	\right. 
	\label{eq:one}
\end{equation}
where $\mathbf{a}_{j}$ is the embedding of word $j$ in the aspect sequence. $\mathbf{a}$ denotes the embedding of the aspect.

\vspace{0.05cm}
\subsection{BERT Encoder} The architecture of BERT~\cite{devlin2019bert} is akin to the Transformer~\cite{vaswani2017attention}. 
For simplicity, we omit some architecture details such as position encoding, layer normalization~\cite{xu2019understanding} and residual connections~\cite{he2016deep}.

\vspace{0.05cm}
\textbf{\emph{1) Multi-head Self-attention Mechanism.}} In recent years, the multi-head self-attention mechanism (MultiHead) has received a wide range of applications in natural language processing. In the paper, we adopt MultiHead with $h$ heads to obtain the overall semantics of the whole sentence. The product from each self-attention network is then concatenated and finally transformed as:
\begin{equation}
    \centering
    \begin{split}
	& \mathbf{m} = \left\{\mathbf{m}_{i} \mid i=1,2, \ldots, l_{s}\right\} \\
	&\ \ \ \ = \textbf{MultiHead}(\mathbf{s}\mathbf{W}^Q_h, \mathbf{s}\mathbf{W}^K_h, \mathbf{s}\mathbf{W}^V_h), \\
    \end{split}
	\label{eq:six}
\end{equation}
where $h$ denotes the $h$-th attention head, $\mathbf{W}^Q_i$, $\mathbf{W}^K_i$ and $\mathbf{W}^V_i$ are learnable parameters. Finally, the output feature is $\mathbf{m} = \left\{\mathbf{m}_{i} \mid i=1,2, \ldots, l_{s}\right\}$. For detailed implementation of \textbf{MultiHead}, please refer to Transformer~\cite{vaswani2017attention}.

\vspace{0.05cm}
\textbf{\emph{2) Position-wise Feed-Forward Network.}} 
Since the multi-head attention is a series of linear transformations, we then apply the position-wise feed-forward network (FFN) to learn the feature's non-linear transformation. Specifically, the FFN consists of two linear transformations along with a ReLU activation in between. More formally:
\begin{equation}
    \centering
    \begin{split}
    &\mathbf{f} = \left\{\mathbf{f}_{i} \mid i=1,2, \ldots, l_{s}\right\} \\
    &\ \ = \textbf{max}(0, \mathbf{m} \mathbf{W}_1 + \mathbf{b}_1)\mathbf{W}_2 + \mathbf{b}_2,\\
    \end{split}
    \label{eq:seven}
\end{equation}
where $\mathbf{W}_1$, $\mathbf{b}_1$, $\mathbf{W}_2$ and $\mathbf{b}_2$ are learnable parameters in the linear transformations. 

So far, with the input ${S}=\{w_1^s, w_2^s, ..., w_{l_s}^s\}$, we obtain the hidden states $\mathbf{f} = \left\{\mathbf{f}_{i} \mid i=1,2, \ldots, l_{s}\right\}$ via the BERT encoder. Then, for the words' hidden states of the sentence from FFN, we utilize the max-pooling operation to fairly select crucial features in the sentence~\cite{lai2015recurrent,zhang2019interactive}, so as to obtain the original sentence representation $\mathbf{h}_s$ at the beginning of each re-weighting step:   
\begin{equation}
	\begin{split}
	& \mathbf{h}_s = \mathrm{Max\_Pooling}(\mathbf{f}_{i} \mid i=1,2, \ldots, l_{s}).
	\end{split}
	\label{eq:eight}
\end{equation}


\subsection{Dynamic Re-weighting Adapter (DRA)} 
The currently attention mechanism in deep learning is essentially similar to the selective visual attention of human beings~\cite{vaswani2017attention,you2016image}. 
However, as for the text semantic understanding, human brain will discover the intentional relationship of words at a sentential level~\cite{taatgen2007integrated,sha2016reading,sen2020human} and link the incoming semantic information with pre-existing information stored within memory. Thus, we design a dynamic re-weighting adapter (DRA) which can dynamically emphasize the important aspect-aware words for the ABSA task. 

As shown in the right part of Figure~\ref{fig:model}, based on overall semantics of the whole sentence, DRA further selects the most important word at each step with consideration of the specific aspect representation. Specifically, the inputs of DRA are the final outputs of the BERT encoder (i.e., $\mathbf{h}_s$) and the original aspect embedding (i.e., $\mathbf{a}$). In each step, we first utilize re-weighting attention to choose the word for current input from the input sequence ($\left\{\mathbf{s}_{i} \mid i=1,2, \ldots, l_{s}\right\}$). Then, we utilize Gated Recurrent Unit (GRU)\cite{cho2014learning} to encode the chosen word and update the semantic representation of the review sentence.

Formally, we regard the calculation process as:
\begin{equation}
\begin{aligned}
\label{con:five}
	{\mathbf{a}}_{t} &=\mathnormal{F}\left(\left[\mathbf{s}_{1}, \mathbf{s}_{2}, \ldots, \mathbf{s}_{l_{s}}\right], {\mathbf{h}}_{t-1}, \mathbf{a}\right) , \\
	{\mathbf{h}}_{t} &=\mathnormal{GRU}\left({\mathbf{a}}_{t}, {\mathbf{h}}_{t-1}\right), \quad t \in [1, T]
\end{aligned}
\end{equation}
where $\mathbf{a}$ is the original embedding vector of the aspect words. ${\mathbf{a}}_{t}$ is the output of re-weighting function $\mathnormal{F}$. $T$ denotes the dynamic re-weighting length over the sentences, which represents the cognitive threshold of human beings. 
$\mathbf{h}_0 = \mathbf{h}_s$ is the initial state and $\mathbf{h}_T$ is the output hidden states of DRA.

\vspace{0.05cm}
\textbf{\emph{1) The Re-weighting Function.}} More specifically, we utilize the attention mechanism to achieve the re-weighting function $\mathrm{F}$, which aims to select the most important aspect-related word at each step. The calculation can be formulated as: 
\begin{equation}
\begin{aligned}
\label{con:six}
	&\ \ \ \ \ \ \ \ \ \ \ \ \ \ \ \ \ \ \  {\mathbf{S}}\ =\left[\mathbf{s}_{1}, \mathbf{s}_{2}, \ldots, \mathbf{s}_{l_{s}}\right], \\
	&\mathbf{M}= \mathbf{W}_{s} \mathbf{S}+\left(\mathbf{W}_{d} {\mathbf{h}}_{t-1}+\mathbf{W}_{a} \mathbf{a}\right) \otimes \mathbf{w}, \\
	&\ \ \ \ \ \ \ \ \ \ \ \ \ \ \ \ \ \  {\mathbf{m}} =\mathbf{\omega}^{T} \tanh \left(\mathbf{M}\right),
\end{aligned}
\end{equation}
where $ {\mathbf{S}}$ denotes the original sentence embedding, $\mathbf{M}$ is the fusion representation of the aspects and the sentences. $\mathbf{W}_s$, $\mathbf{W}_d$, $\mathbf{W}_a$ and $\omega$ are trainable parameters. $\mathbf{w} \in \mathbb{R}^{l_s}$ is a row vector of 1 and $\otimes$ denotes the outer product.

Subsequently, to better encode aspect-aware semantics, we choose the most important word (i.e., one word) at each step for the specific aspect. 
\begin{equation}
\begin{aligned}
\label{con:sev}
	& \alpha_i =  \frac{\exp \left( {m}_{i}  \right)}{\sum_{k=1}^{l_{s}} \exp \left( {m}_{k}\right)}\ , \\
	&{\mathbf{a}_t} =\mathbf{s}_j, (j = \operatorname{Index}(\operatorname{max}(\alpha_i )))
\end{aligned}
\end{equation}
where $m_i$ and ${\alpha}_i$ are the hidden state and the attention score of $i$-th word in the sentence. ${\mathbf{a}}_{t}$ is the chosen word which is most related to the specific aspect at $t$-th step. However, Index(max($\cdot$)) operation has no derivative, which means its gradient could not be calculated. Inspired by softmax function, we modify the Eq.\ref{con:sev} and employ the following operation to re-weight the contextual words:
\begin{equation}
\begin{split}
	&\ {\mathbf{a}_t} =\sum_{i=1}^{l_{s}} \frac{\exp \left(\lambda {m}_{i}  \right)}{\sum_{k=1}^{l_{s}} \exp \left(\lambda {m}_{k}\right)}  \mathbf{s}_i\ .
	\label{eq:aaa}
\end{split}
\end{equation}

Note that, we design a hyper-parameter $\lambda$ to ensure our model achieves the above purpose. Specifically,  the softmax function can exponentially increase or decrease the signal, thereby highlighting the information we want to enhance. Thus, when $\lambda$ is an arbitrarily large value, the attention score of the chosen word is infinitely close to 1, and other words are infinitely close to 0. In this way, the most important word (i.e., one word) will be extract from the context at each re-weighting step.

\vspace{0.05cm}
\textbf{\emph{2) The GRU Function.}} To better encode semantic of the whole sentence, we also employ GRU to further imitate the procedure of human semantic comprehension under the specific context, which is consistent with the process of people adjusting to a new text based on their understanding behavior. Therefore, given a previous vector embedding, the hidden vectors of GRU are calculated by receiving it as input:
\begin{equation}
\begin{split}
	& {z}_{t} =\sigma\left(\mathbf{W}_{z} \cdot\left[\mathbf{h}_{t-1}, \mathbf{a}_{t}\right]\right) \\
	& {r}_{t} =\sigma\left(\mathbf{W}_{r} \cdot\left[\mathbf{h}_{t-1}, \mathbf{a}_{t}\right]\right) \\
	& \tilde{\mathbf{h}}_{t} =\tanh \left(\mathbf{W} \cdot\left[r_{t} * \mathbf{h}_{t-1}, \mathbf{a}_{t}\right]\right) \\
	& \mathbf{h}_{t} =\left(1-z_{t}\right) * \mathbf{h}_{t-1}+z_{t} * \tilde{\mathbf{h}}_{t}\ , 
\end{split}
\end{equation}
where $\sigma$ is the logistic sigmoid function. $z_t$ and $r_t$ denote the update gate and reset gate respectively at the time step $t$. 

\subsection{Sentiment Predicting}
After applying BERT layers and DRA on the input sentence, its root representation (i.e., $\mathbf{s}$) is convert into the feature representation $\mathbf{e}$:
\begin{equation}
    \centering
    \begin{split}
    &\mathbf{e} = \left\{\mathbf{e}_{i} \mid i=1,2, \ldots, l_{s}\right\} \\
    &\ \ = \left(\mathbf{W}_{e} \mathbf{f} +\mathbf{U}_{e}{\mathbf{h}_T} + \mathbf{b}_{e}\right),\\
    \end{split}
\end{equation}
where $\mathbf{W}_{e}$, $\mathbf{U}_{e}$ and $\mathbf{b}_{e}$ are trainable parameters. After $N$-th stacked BERT layers, we obtain the final representation of the sentence (i.e., $\mathbf{e}_N$). Then, we feed it into a Multilayer Perceptron (MLP) and map it to the probabilities over the different sentiment polarities via a softmax layer:
\begin{equation}
\begin{split}
	& \mathbf{R}_{l}=\mathrm{Relu} (\mathbf{W}_{l} {\mathbf{R}_{l-1}}+\mathbf{b}_{l})\ , \\
	& \hat{\mathbf{y}}=\mathrm{softmax}\left(\mathbf{W}_{o}{\mathbf{R}_h}+\mathbf{b}_{o}\right),
	\end{split}
\end{equation}
where $\mathbf{W}_{l}$, $\mathbf{W}_{o}$ , $\mathbf{b}_{l}$ and $\mathbf{b}_{o}$ are learned parameters. $\mathbf{R}_l$ is the hidden state of $l$-th layer MLP ($\mathbf{R}_0 = \mathbf{e}_N$, $l \in [1, h]$). $\mathbf{R}_h$ is the state of final layer which is also regard as the output of the MLP. $\hat{\mathbf{y}}$ is the predicted sentiment polarity distribution.

\subsection{Model Training}
Finally, we applies the cross-entropy loss function for model training:
\begin{equation}
\mathcal{L}=-\sum_{i=1}^{M} \sum_{j=1}^{C} y_{i}^{j} \log \left(\hat{y}_{i}^{j}\right) + \beta\|\Theta\|_{2}^{2}\ ,
\vspace{-0.2cm}
\end{equation}
where ${y_i^j}$ is the ground truth sentiment polarity. $C$ is the number of labels (i.e, 3 in our task). $M$ is the number of training samples. $\Theta$ corresponds to all of the trainable parameters. 

%

\begin{table} 
    \vspace{0.18cm}
    \small
    \centering
    \begin{tabular}{p{1.1cm}p{0.35cm}<{\centering}p{0.25cm}<{\centering}p{0.35cm}<{\centering}p{0.25cm}<{\centering}p{0.35cm}<{\centering}p{0.25cm}<{\centering}p{0.05cm}<{\centering}p{0.35cm}<{\centering}}
            \toprule
            \multirow{2.5}{*}{Datasets} &\multicolumn{2}{c}{\#Positive}&\multicolumn{2}{c}{\#Negative}&\multicolumn{2}{c}{\#Neural}&\multirow{2}{*}{\#L}&\multirow{2}{*}{\#M}\\
            \specialrule{0em}{2pt}{0pt}
            \cline{2-3}
            \cline{4-7}
            \specialrule{0em}{3pt}{0pt}
            & Train & Test & Train & Test & Train & Test \\
            \midrule
            Restaurant &2164&728&807&196&637&196&20&45.5\\
            \specialrule{0em}{2pt}{0pt}
            Laptop 
&994&341&870&128&464&169&19&36.5\\
            \specialrule{0em}{2pt}{0pt}
            Twitter &1561&173&1560&173&3127&346&16&10.2\\
            \bottomrule
    \end{tabular}
    \caption{The statistics of three benchmark datasets. \#L is the average length of sentences. \#M is the proportion (\%) of samples with multiple (i.e., more than 1) aspects.} 
    \label{tab:one} 
    \vspace{-0.3cm}
\end{table}

\begin{table*} 
    \centering
    \small
    \subtable{
        \begin{tabular}{p{1.1cm}p{4.5cm}p{1.1cm}<{\centering}p{1.2cm}<{\centering}p{1.1cm}<{\centering}p{1.2cm}<{\centering}p{1.1cm}<{\centering}p{1.2cm}<{\centering}}
            \toprule
            \multirow{2}{*}{Category} &\multirow{2.5}{*}{\diagbox [width=11em,trim=l] {\ \ Methods}{Datasets}} &\multicolumn{2}{c}{Laptop}&\multicolumn{2}{c}{Restaurant}&\multicolumn{2}{c}{Twitter} \\
            \specialrule{0em}{2pt}{0pt}
            \cline{3-8}
            \specialrule{0em}{3pt}{0pt}
            & &Accuracy&F1-score&Accuracy&F1-score&Accuracy&F1-score\\
            \specialrule{0em}{2pt}{0pt}
            \midrule 
            \multirow{7}{*}{Attention.}
            &\ \ ATAE-LSTM~\cite{wang2016attention} 
            &68.57
            &64.52
            &76.58
            &67.39
            &67.27
            &66.43
            \\
            \specialrule{0em}{1pt}{0pt}
            &\ \ IAN~\cite{ma2017interactive} 
            &70.84
            &65.73
            &76.88 
            &68.36
            &68.74
            &67.61
            \\
            \specialrule{0em}{1pt}{0pt}
            &\ \ MemNet~\cite{tang2016aspect} 
            &72.32
            &67.03
            &78.12 
            &68.99
            &70.19
            &68.22
            \\
            \specialrule{0em}{1pt}{0pt}
            &\ \ AOA~\cite{huang2018aspect} 
            &74.56
            &68.77
            &79.42 
            &70.43
            &71.68
            &69.25
            \\
            \specialrule{0em}{1pt}{0pt}
            &\ \ MGNet~\cite{fan2018multi} 
            &75.37
            &71.26
            &81.28 
            &72.07
            &72.54
            &70.78
            \\
            \specialrule{0em}{1pt}{0pt}
            &\ \ TNet~\cite{li2018transformation}
            &76.54
            &71.75
            &80.69 
            &71.27
            &74.93
            &73.60
            \\
            \midrule
            \specialrule{0em}{1pt}{0pt}
            \midrule
            \specialrule{0em}{2pt}{0pt}
            \multirow{6.5}{*}{Pre-trained.}            
            &\ \ BERT~\cite{devlin2019bert} 
            &{77.29}
            &{73.36}
            &{82.40}
            &{73.17}
            &{73.42}
            &{72.17}
            \\
            \specialrule{0em}{1pt}{0pt}
            &\ \  BERT-PT~\cite{xu2019bert}
            &{78.07}
            &{75.08}
            &{84.95}
            &{76.96}
            &{--}
            &{--}
            \\
             \specialrule{0em}{1pt}{0pt}
            &\ \ BERT-SPC~\cite{song2019attentional}
            &{78.99}
            &{75.03}
            &{84.46} 
            &{76.98}
            &{74.13}
            &{72.73}
            \\
            \specialrule{0em}{1pt}{0pt}
            &\ \ AEN-BERT~\cite{song2019attentional} 
            &{79.93}
            &{76.31}
            &{83.12} 
            &{73.76}
            &{74.71}
            &{73.13}
            \\
            \specialrule{0em}{1pt}{0pt}
            &\ \ RGAT-BERT~\cite{wang2020relational}
            &{78.21}
            &{74.07}
            &\underline{86.60}
            &\underline{81.35}
            &{76.15}
            &{74.88}
            \\
            \specialrule{0em}{1pt}{0pt}
            &\ \ T-GCN~\cite{tian2021aspect}
            &\underline{80.88}
            &\underline{77.03}
            &{86.16}
            &{79.95}
            &\underline{76.45}
            &\underline{75.25}
            \\
            \midrule
            \specialrule{0em}{1pt}{0pt}
            \midrule
            \multirow{1}{*}{\textbf{Ours.}}
            &\ \ \textbf{DR-BERT} 
            &\textbf{81.45}
            &\textbf{78.16}
            &\textbf{87.72} 
            &\textbf{82.31}
            &\textbf{77.24}
            &\textbf{76.10}            
            \\
            \bottomrule
        \end{tabular}
    } 
    \vspace{-0.2cm}
        \caption{Experimental results (\%) in three benchmark datasets. We underline the best performed baseline.} 
        \label{tab:two}
        \vspace{-0.1cm}
\end{table*}

\section{Experiment}

\subsection{{Datasets}} We mainly conduct experiments on three benchmark ABSA datasets, including ``Laptop'', ``Restaurant''~\cite{pontiki-etal-2014-semeval} and ``Twitter''~\cite{dong2014adaptive}. Each data item is labeled with three sentiment polarities (i.e., {positive}, {negative} and {neutral}). The statistics of the datasets are presented in Table~\ref{tab:one}.
Moreover, we follow the dataset configurations of previous studies strictly. For all datasets, we randomly sample 10\% items from the training set and regard them as the development set.

\subsection{Hyperparameters Settings} 
In the implementation, we build our framework based on the official {bert-base} models (n$_{\operatorname{layers}}$=12, n$_{\operatorname{heads}}$=12, n$_{\operatorname{hidden}}$=768). The hidden size of GRUs and re-weighting length of DRA are set to 256 and 7. The learning rate is tuned amongst [2e-5, 5e-5 and 1e-3] and the batch size is manually tested in [16, 32, 64, 128]. The dropout rate is set to 0.2. The hyper-parameter $l$ , $\beta$ and $\lambda$ have been carefully adjusted, and final values are set to 3, 0.8 and 100 respectively. The model is trained using the Adam optimizer and evaluated by two widely used metrics. The parameters of baseline models are in accordance with the default configuration of the original paper.  We run our model three times with different seeds and report the average performance. 

\subsection{{Baselines}} 
\begin{itemize}
	\item \textbf{{Attention-based Models:}} MemNet~\cite{tang2016aspect},
ATAE-LSTM~\cite{wang2016attention}, IAN~\cite{ma2017interactive}, AOA~\cite{huang2018aspect}, MGNet~\cite{fan2018multi}, TNet~\cite{li2018transformation}.
\vspace{-0.2cm}
	\item \textbf{{Pre-trained Models:}} Fine-tune BERT~\cite{devlin2019bert}, BERT-PT~\cite{xu2019bert}, BERT-SPC, AEN-BERT~\cite{song2019attentional}, RGAT-BERT~\cite{wang2020relational}, T-GCN~\cite{tian2021aspect}. 
\end{itemize}

The baseline methods have comprehensive coverage of the recent related SOTA models recently. Most of them are detailed in Section~\ref{absa}. For space-saving, we do not detail them in this section.

\subsection{Experimental Results}
From the results in Table~\ref{tab:two}, we have the following observations. First, BERT-based methods beat most of the attention-based methods (e.g., IAN and TNet) in both metrics. 
The phenomenon indicates the powerful ability of the pre-trained language models. That is also why we adopt BERT as base encoder to learn the overall semantic representation of the whole sentences.

Second, by comparing non-specific BERT models (i.e., BERT and BERT-PT) with task-specific models (e.g., RGAT-BERT) for ABSA, we find that the task-specific BERT models perform better than the non-specific models. 
Specifically, we can also observe the performance trend that T-GCN\&RGAT-BERT $>$AEN-BERT$>$BERT-PT$>$BERT, which is consistent with the previous assumption that aspect-related information is the crucial influence factor for the performance of the ABSA model.

Finally, despite the outstanding performance of previous models, our DR-BERT still outperforms the most advanced baseline (i.e., T-GCN or RGAT-BERT) no matter in terms of Accuracy or F1-score. The results demonstrate the effectiveness of the dynamic modeling strategy based on the procedure of semantic comprehension. Meantime, it also indicates that our proposed DRA can better grasp the aspect-aware semantics of the sentence than other BERT plus-in components in previous methods.

\subsection{Ablation Study} 
\label{ablation}

\begin{table}
    \vspace{-0.05cm}
    \small
    \centering
     \begin{tabular}{p{3.6cm}|p{1.35cm}<{\centering}p{1.15cm}<{\centering}}       
            \midrule
            \specialrule{0em}{3pt}{0pt}
            \multirow{2.5}{*}{Model Variants}&\multicolumn{2}{c}{Laptop}\\ 
             \specialrule{0em}{2pt}{0pt}
            \cline{2-3}
            \specialrule{0em}{3pt}{0pt}
             &{\ \ Accuracy}&{F1-score}\\
            \midrule 
            \specialrule{0em}{3pt}{0pt} 
            \ \ \ \ \ \ \ BERT-Base &{\ \ 77.29}& {73.36}\\
            \specialrule{0em}{3pt}{0pt}
            (1): + MLP &\ \ 77.94&74.42\\ 
            \specialrule{0em}{3pt}{0pt}
            (2): + DRA &\ \ 80.66&77.13\\ 
            \specialrule{0em}{3pt}{0pt}
            \midrule
            \specialrule{0em}{1pt}{0pt}
            \midrule
            (3): + DRA on top 3 layers &\ \ 78.64&75.16\\ 
            \specialrule{0em}{2pt}{0pt}
            (4): + DRA on top 6 layers &\ \ 79.17&75.93\\ 
            \specialrule{0em}{2pt}{0pt}
            (5): + DRA on top 9 layers &\ \ 80.22&76.49\\ 
            \specialrule{0em}{2pt}{0pt}
            \textbf{(6): DR-BERT} &\ \ \textbf{81.45}&\textbf{78.16}\\ 
            \midrule
    \end{tabular}
    \vspace{-0.15cm}
    \caption{{The ablation study on different components which conducted on the test set of the Laptop dataset. ``BERT-Base'' indicates the vanilla BERT. ``+'' indicates the setting with plus-in components.}} 
    \label{tab:ablation}
    \vspace{-0.05cm}
\end{table}

\begin{figure}
	\centering
	\vspace{-0.1cm}
	\includegraphics[width=1\columnwidth]{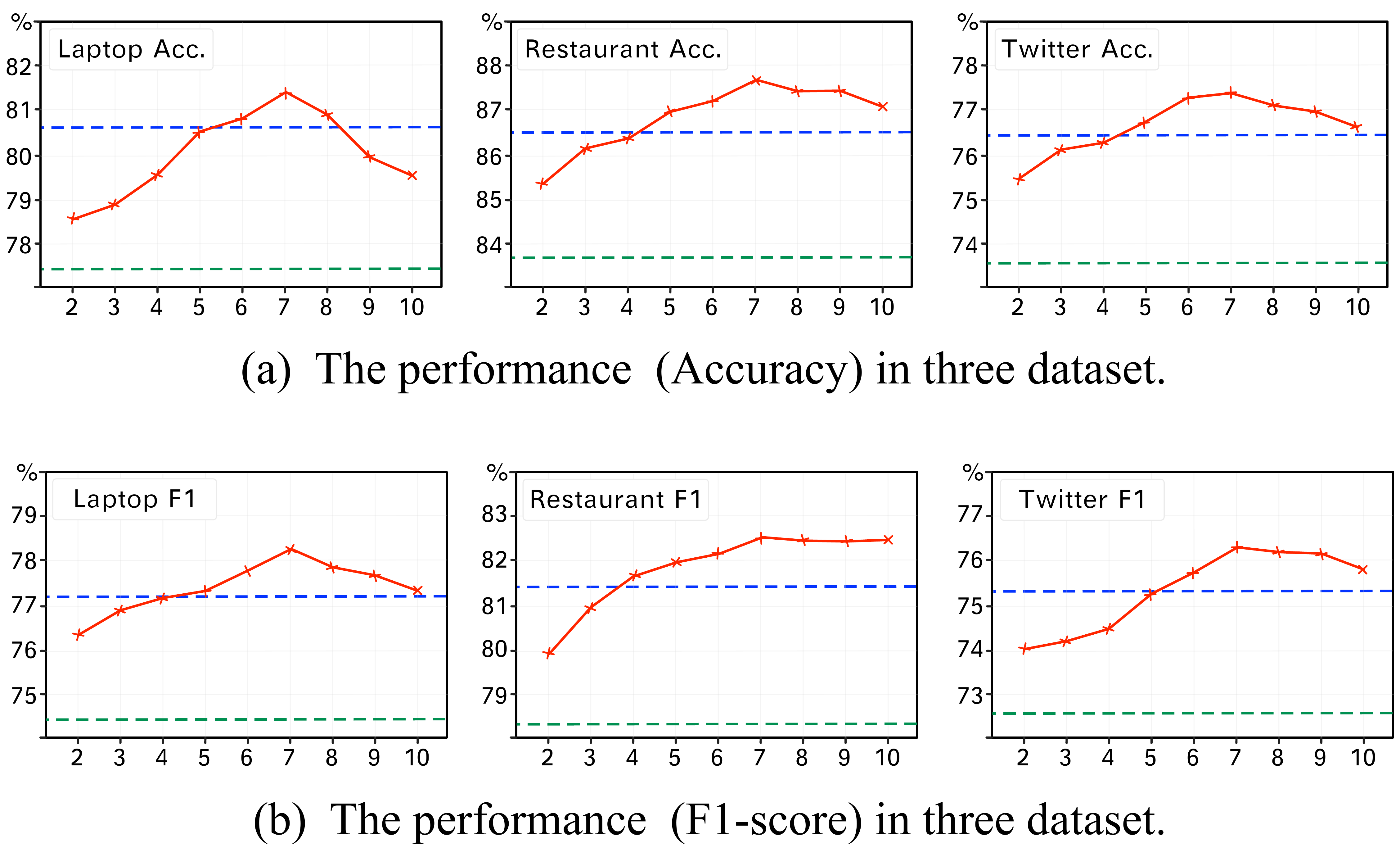}
	\vspace{-0.6cm}
	\caption{The ablation study on the re-weighting length of the adapter. Red lines indicate Accuracy/ F1 scores while blue and green lines indicate the performance of the best baseline and BERT-base model respectively.
	}
	\label{fig:para}
	\vspace{-0.45cm}
\end{figure}

\vspace{-0.05cm}
\textbf{\ \ \ \ Ablations on the Proposed Components.}
In Table~\ref{tab:ablation}, we study the influence of different components in our framework, including the DRA and MLPs. We can find that without utilizing adapters and MLPs, DR-BERT degenerates into the BERT model, which gains the worst performance among all the variants. The phenomenon indicates the effective of the DRA and MLP modules. Moreover, through comparing (1) and (2), we can easily conclude that DRA plays a more crucial role in the final sentiment prediction than MLPs. 

Since BERT models are usually quite deep (e.g., 12 layers), we only insert the dynamic re-weighting adapter into top layers (i.e., 3-th, 6-th, and 9-th layers) to further verify the effectiveness of the DRA module. The results are shown in Table~\ref{tab:ablation} (3), (4), and (5). We observe that when introducing adapters to the top layers of DR-BERT, our framework still outperforms the BERT model, showing that the DRA is efficient in encoding the aspect-aware semantics over the whole sentence. In addition, we can also find that the more adapter incorporated in BERT layers the higher performance gained, illustrating the importance of modeling the deep dynamic semantics over the sentence.

\vspace{0.05cm}
\textbf{Ablations on the Scale of Adapter.} 
In this subsection, we investigate the influence of the scale of adapters on different datasets. As shown in Figure~\ref{fig:para}, we tune the adapter's dynamic re-weighting length ($T$) in a wide range (i.e., 2 to 10). 
Specifically, the performance of DR-BERT first becomes better with the increasing of re-weighting length and achieving the best result at around 7. Then, as the length continues to increase, the performance continues to decline. This phenomenon is consistent with the psychological findings that human memory focuses on nearly seven words~\cite{tononi2008consciousness,koch2007attention}, which further indicates the effectiveness of DRA in modeling human-like (dynamic) semantic comprehension. 

Besides, compared with the best-performed baseline (blue lines), our model can achieve better performance with only 4 or 5 times of re-weighting at most test sets, illustrating the efficiency of the re-weighting adapter. On the other hand, we can also find that DR-BERT always gives superior performance compared to the BERT-based model (green lines), even with the lowest re-weighting length. All those results show that DR-BERT could better comprehend aspect-aware dynamic semantics in aspect-based sentiment analysis. 

\begin{figure}
	\centering
	\includegraphics[width=0.75\columnwidth]{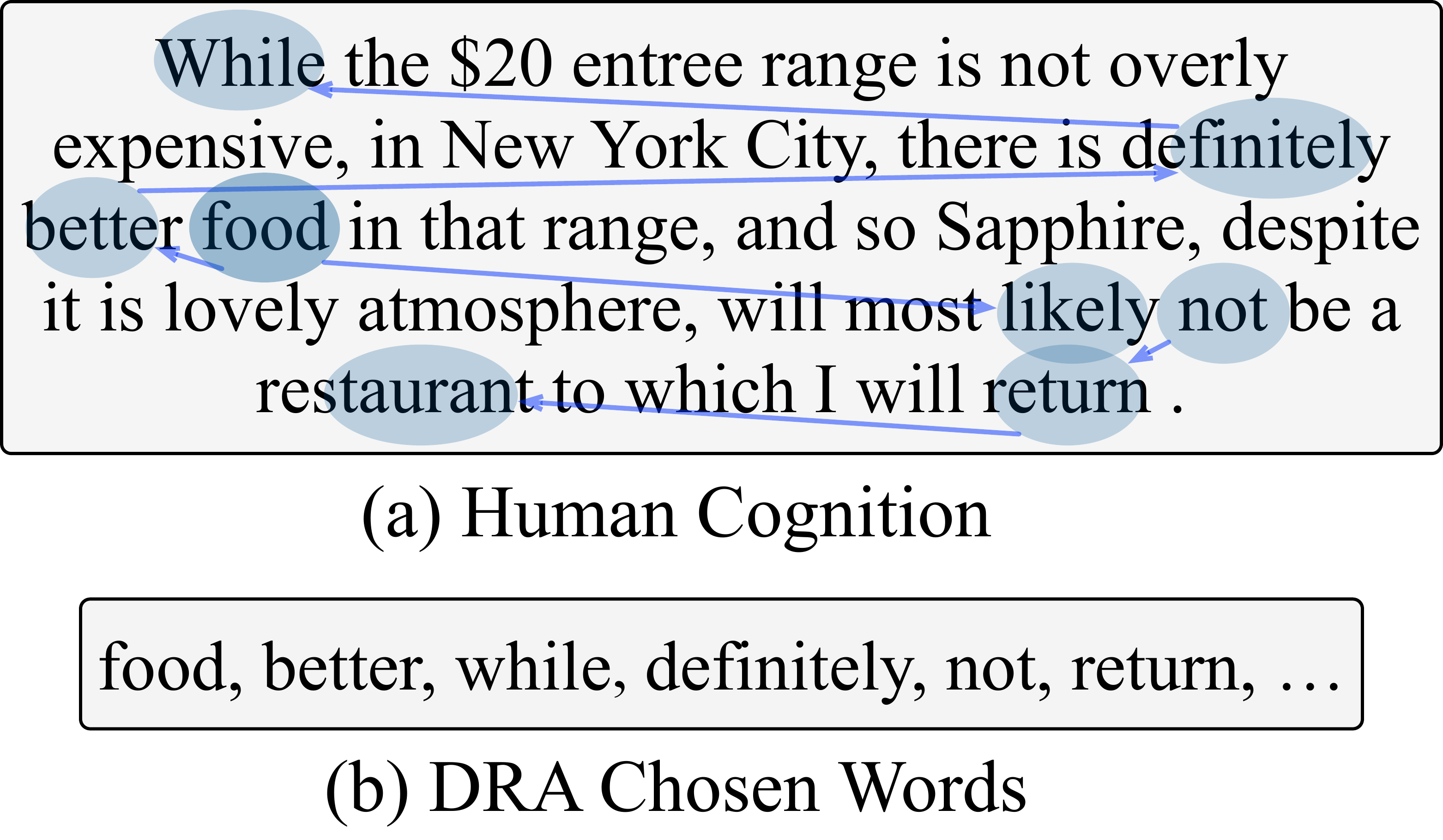}
	\vspace{-0.2cm}
	\caption{Comparison of the semantic understanding process between human reading and DRA when judging the sentiment polarity of aspect \emph{``food''}. (a) is the visualization of the human understanding process from the eye tracker\protect\footnotemark[2]. (b) denotes aspect-aware words from re-weighting function.
	}
	\label{fig:case1}
	\vspace{-0.15cm}
\end{figure}
\footnotetext[2]{The procedure of the human semantic comprehension is generated by the eye tracker: \url{https://www.tobiipro.com/product-listing/nano/}}

\begin{figure*}
	\centering
	\includegraphics[width=1.88\columnwidth]{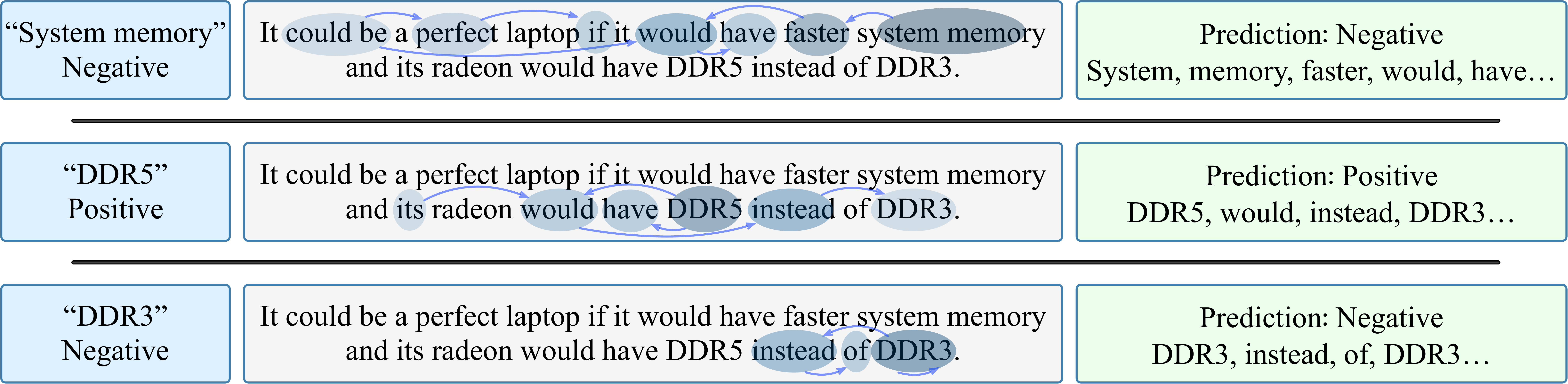}
	\caption{Visualization results of multiple aspects in the same sentence. The blue part indicates the aspect and its ground truth. The middle subfigures represent the procedure of human's semantic comprehension which is targeted at one specific aspect. The green subfigures are the predicted labels and the chosen word sequences from DRA.
	}
	\label{fig:case2}
\end{figure*}

\begin{table*} 
    \centering
    \small
    \begin{tabular}{p{9.3cm}|p{1.65cm}|p{1.7cm}|p{1.5cm}}
            \toprule
            \specialrule{0em}{3pt}{0pt}
            {\textbf{Case Examples}. The label in brackets represents ground truth.} &{BERT-base} & {RGAT-BERT} & {DR-BERT}\\ 
            \midrule
             \textbf{Aspects}:  ``system memory''(\textcolor[rgb]{0.8,0,0}{{Neg.}}), ``DDR5''(\textcolor[rgb]{0.5,0.5,0}{{Pos.}}), ``DDR3''(\textcolor[rgb]{0.8,0,0}{{Neg.}}) 
             &Pos/Neg/Neg
             &Neg/Pos/Pos
             &\textcolor[rgb]{0,0,0}{{Neg}}/\textcolor[rgb]{0,0,0}{{Pos}}/\textcolor[rgb]{0,0,0}{{Neg}}\\
             \textbf{Sentence}: It could be a perfect laptop if it would have faster system memory and its radeon would have DDR5 instead of DDR3. 
             &\multirow{2}{*}{\XSolidBrush\ / \ \XSolidBrush\ / \XSolidBrush} 
             &\multirow{2}{*}{\CheckmarkBold\ / \ \CheckmarkBold\ / \XSolidBrush} 
             & \multirow{2}{*}{
             \textcolor[rgb]{0,0,0}{\CheckmarkBold}\ 
             / \textcolor[rgb]{0,0,0}{\CheckmarkBold}\ 
             / \textcolor[rgb]{0,0,0}{ \CheckmarkBold}}\\
            \midrule
            \specialrule{0em}{3pt}{0pt}
             \textbf{Aspects}:  ``Supplied software'' (\textcolor[rgb]{0.5,0.5,0}{{Neu.}}), ``software'' (\textcolor[rgb]{0,0.77,0}{{Pos.}}), ``Windows'' (\textcolor[rgb]{0.8,0,0}{Neg.})
             &Pos/ Pos/ Pos
             &Pos/Pos/Neu
             &Pos/Pos/Neg\\         
             \textbf{Sentence}: Supplied software: The software that comes with this machine is greatly welcomed compared to what Windows comes with.
             &\multirow{2}{*}{\XSolidBrush\ / \ \CheckmarkBold\ / \XSolidBrush} 
             &\multirow{2}{*}{\XSolidBrush\ / \ \CheckmarkBold\ / \XSolidBrush} 
             & \multirow{2}{*}{\textcolor[rgb]{0,0,0}{\XSolidBrush}\ / \textcolor[rgb]{0,0,0}{\CheckmarkBold}\ / \textcolor[rgb]{0,0,0}{ \CheckmarkBold}}\\
             \midrule
            \specialrule{0em}{3pt}{0pt}
             \textbf{Aspects}:  ``waiter'' (\textcolor[rgb]{0.8,0,0}{Neg.}), ``served'' (\textcolor[rgb]{0.8,0,0}{Neg.}), ``specials'' (\textcolor[rgb]{0,0.77,0}{Pos.})
             &Neg/Neg/Neg
             &Neg/Neg/Neu
             &Neg/Neg/Pos\\         
             \textbf{Sentence}: First, the waiter who served us neglected to fill us in on the specials, which I would have chosen had I known about them.
             &\multirow{2}{*}{\CheckmarkBold\ / \ \CheckmarkBold\ / \XSolidBrush} 
             &\multirow{2}{*}{\CheckmarkBold\ / \ \CheckmarkBold\ / \XSolidBrush} 
             & \multirow{2}{*}{\textcolor[rgb]{0,0,0}{\CheckmarkBold}\ / \textcolor[rgb]{0,0,0}{\CheckmarkBold}\ / \textcolor[rgb]{0,0,0}{ \CheckmarkBold}}\\
           \bottomrule
    \end{tabular}
    \vspace{-0.05cm}
        \caption{{Error analysis of two review items from laptop and restaurant. The colored words in brackets represents ground truth sentiment label of the corresponding aspect. The symbol $\checkmark$ means the predicting sentiment is correct, and the other symbol means the predicting sentiment is wrong.}} 
    \label{tab:case} 
\end{table*} 

\subsection{Interpretability Verification}
\textbf{\ \ \ \ Comparison of Semantic Comprehension.} To evaluate model rationality and interpretability, we conduct an study for dynamic semantic comprehension by eye tracker. As shown in Figure~\ref{fig:case1} (a), when a person tries to understand a relatively long sentence, he/she first read the entire sentence. 
Subsequently, after giving a specific aspect, he/she will dynamically select related words based on the previous memory state until he/she fully understands the sentiment polarity of the given aspect.

Interestingly, the above phenomenon is consistent with our dynamic re-weighting adapter's chosen result. Specifically, as Figure~\ref{fig:case1} (b) shows, with the re-weighting function $F$ (i.e., Equation~\ref{con:five} and~\ref{con:six}), our model dynamically choose the words \emph{``food, better, while, definitely, not, ...''}, which have proven to be very important for predicting the sentiment of aspect \emph{``food''} in Figure~\ref{fig:case1} (a). Those experimental results again fully indicate the effectiveness and interpretability of our proposed model in dynamic learning aspect-aware information.

\begin{table*}
    \small
    \centering
    \label{tab:time} 
    \begin{tabular}{p{3.2cm}<{\centering}|p{0.8cm}<{\centering}p{0.88cm}<{\centering}p{1cm}<{\centering}|p{0.8cm}<{\centering}p{0.88cm}<{\centering}p{1cm}<{\centering}|p{0.8cm}<{\centering}p{0.88cm}<{\centering}p{1cm}<{\centering}}       
            \bottomrule
            \specialrule{0em}{3pt}{0pt}
            \multirow{2.5}{*}{\ \ \ \ \ \textbf{Methods}}&\multicolumn{3}{c}{\textbf{Laptop}} &\multicolumn{3}{c}{\textbf{Restaurant}} &\multicolumn{3}{c}{\textbf{Twitter}} \\ 
            \specialrule{0em}{2pt}{0pt}
            \cline{2-10}
            \specialrule{0em}{2pt}{0pt}
            &\textbf{S}&\textbf{E}&\textbf{T}&\textbf{S}&\textbf{E}&\textbf{T}&\textbf{S}&\textbf{E}&\textbf{T}\\ 
            \midrule 
            \specialrule{0em}{3pt}{0pt} 
            (1) DR-BERT &{157s}& {10}&{26.1m}&{183s}&{10}&{30.5m}&{379s}&{10}&63.2m\\
            \specialrule{0em}{3pt}{0pt}
            (2) T-GCN-BERT &168s&10&28.0m&188s&10&31.3m&411s&10&68.5m\\ 
            \specialrule{0em}{3pt}{0pt}
            (3) BERT-base &133s&10&22.2m&158s&10&26.3m&242s&10&40.3m\\ 
            \specialrule{0em}{3pt}{0pt}
            (4) ATAE-LSTM &3s&30&1.50m&4s&30&2.00m&5s&30&2.50m\\ 
            \specialrule{0em}{3pt}{0pt}
            \bottomrule
    \end{tabular}
        \caption{{Runtime comparison between DR-BERT, T-GCN-BERT, BERT-base and ATAE-LSTM. Specifically, ``S'' represents the training time (seconds) for a single epoch, ``E'' denotes the number of training epochs, and ``T'' is the total training time (minutes).}} 
    \vspace{-0.15cm}
    \label{tab:time} 
\end{table*}

\vspace{0.05cm}
\textbf{The Influence of multiple Aspects.} As aspect-related information plays a key role in ABSA and at least 10.2\% of reviews contain multiple aspects as shown in Table~\ref{tab:one}, we are curious about the model's performance in the complex scenarios, e.g., a review sentence contains multiple aspects. Therefore, we randomly choose an example to explore how the selection of the context words will correspondingly change with different inputs. The visualization results are shown in Figure~\ref{fig:case2}. Specifically, the chosen sentence has three different aspects with their sentiment polarity, i.e., \emph{``System memory''}-negative, \emph{``DDR5''}-positive and \emph{``DDR3''}-negative. 
Take the aspect \emph{``DDR5''} as example, it is positive which is contrary to \emph{``DDR3''}. After receiving the overall semantic of the whole sentence, readers tend to associate \emph{``DDR5''} with the context words \{\emph{``would'', ``have''}\} to predict the correct sentiment ``positive''. For other two aspects, the observations are consistent with \emph{``DDR5''}. In summary, all those results show that DR-BERT could dynamically extract the vital information to achieve aspect-aware semantic understanding even in a more complex scenario.

\subsection{Error Analysis}
Table~\ref{tab:case} displays three review examples and their prediction results by BERT, RGAT-BERT, and our DR-BERT. As we can see from the ``BERT-base'' column, when there are multiple aspects, the vanilla BERT often makes the wrong classification since it tends to learn the overall sentiment polarity of the sentences instead of the aspect-aware semantic. While RGAT-BERT can alleviate the problem to a certain extent, it is also hard to predict the accurate sentiment label with few dependency relations. For example, in the first sentence, \emph{``DDR3''} has few helpful syntactic dependency relations. Therefore, RGAT-BERT makes a wrong sentiment prediction. However, our DR-BERT model, succeeding in predicting most sentiment labels by considering the dynamic changing of the aspect-aware semantic. For other two case examples, the observations are consistent. Note that, for aspect \emph{``Supplied software''} in second sentence, two overlap aspects appear in the same sentence makes it more difficult to distinguish the different sentiment between them. Thus, precisely determine its sentiment polarity is a big challenge for human, let alone deep learning models. This also leaves space for future exploration.

\section{Computation Time Comparison}
We also compared the computation runtime of three baseline methods. All of the models are performed on a Linux server with 64 Intel(R) CPUs and 4 Tesla V100 32GB GPUs. From the results shown in Table~\ref{tab:time}, we can first observe that the training time of a single epoch in DR-BERT performs better than T-GCN, which is based on GCN. Meanwhile, the training time of all these BERT-based models is similar (i.e., there is no significant difference). The possible reason is that the official datasets are small, and it is hard to influence the overall runtime of PLMs with such a small amount of data. 
Second, compared with other models, the training time of the ATAE-LSTM model is less (always an order of magnitude lower). For example, the ATAE-LSTM only needs about two minutes to achieve optimal performance in the restaurant dataset, while BERT-based models require more than 26 minutes. Therefore, though DR-BERT contains a Dynamic Re-weighting adapter based on GRU, the computation time is much lower than the BERT-based framework. In summary, the observations above show that the computation time of our DR-BERT model is within an acceptable range.

\section{Conclusion and Future Works}
This paper introduced a new approach named Dynamic Re-weighting BERT (DR-BERT) for aspect-based sentiment analysis. Specifically, we first employed the BERT layers as a base encoder to learn the overall semantic features of the whole sentence. Then, inspired by human semantic comprehension, we devised a new Dynamic Re-weighting Adapter (DRA) to enhance aspect-aware semantic features in the sentiment learning process. In addition, we inserted the DRA into the BERT layers to address the limitations of the vanilla pre-trained model in ABSA task. Extensive experiments on three benchmark datasets demonstrated the effectiveness and interpretability of the proposed model, with good semantic comprehension insights for future nature language modeling. Moreover, the error analysis was performed on incorrectly predicted examples, leading to some insights into the ABSA task. 

We hope our research can help boost excellent work for aspect-based sentiment analysis from different perspectives. In the future, we plan to extend our method to other tasks like Sentence Semantic Matching, Relation Extraction, etc., which can also benefit from utilizing the dynamic semantics. Besides, we will explore whether DR-BERT can make any positive changes based on previous mistakes during the dynamic semantic understanding.

\section{Acknowledgments}
We would like to thank the anonymous reviewers for the helpful comments. This research was partially supported by grants from the National Key R\&D Program of China (No. 2021YFF0901003), and the National Natural Science Foundation of China (No. 61922073, 61727809, 62006066 and 72101176). We appreciate all the authors for their fruitful discussions. We also special thanks to all the first-line healthcare providers that are fighting the war of COVID-19. 

\newpage
\bibliography{anthology,custom}
\bibliographystyle{acl_natbib}

\end{document}